\documentclass[letterpaper, 10 pt, journal, twoside]{IEEEtran}
\ifCLASSINFOpdf
\else
\fi
\usepackage{graphics} 
\usepackage{epsfig} 
\usepackage{booktabs}
\usepackage{multirow}
\usepackage{amsmath} 
\usepackage{amssymb}  
\usepackage{color}
\usepackage{hyperref}
\usepackage{cleveref}
\usepackage{url}
\usepackage{float}
\usepackage{siunitx}
\usepackage{todonotes}
\usepackage{layouts}
\usepackage{subcaption}
\usepackage{xcolor}

\usepackage{adjustbox}
\usepackage{array}
\newcolumntype{R}[2]{%
    >{\adjustbox{angle=#1,lap=\width-(#2)}\bgroup}%
    l%
    <{\egroup}%
}
\newcommand*\rot{\multicolumn{1}{R{45}{1em}}}

\usepackage{pifont}
\newcommand{\cmark}{\ding{51}}%
\newcommand{\xmark}{\ding{55}}%

\newcommand{\lightgray}[1]{\textcolor{gray}{#1}}

\newcommand{\rev}[1]{{\color{black} #1}}
\newcommand{\revtwo}[1]{{\color{black} #1}} \newcommand{\revthree}[1]{{\color{black} #1}} 

\newcommand{\norm}[1]{\left\Vert #1 \right\Vert}
\newcommand{\normed}[1]{\frac{#1}{\norm{#1}}}

\DeclareMathOperator*{\argmin}{arg\,min}

\begin{document}
%
\title{Exploring Event Camera-based Odometry for Planetary Robots}
%
%
%

\author{Florian Mahlknecht$^{1}$, Daniel Gehrig$^{2}$, Jeremy Nash$^{1}$, Friedrich M. Rockenbauer$^{1}$, Benjamin Morrell$^{1}$, \\Jeff Delaune$^{1}$, and Davide Scaramuzza$^{2}$
\thanks{Manuscript received: February, 24, 2022; Revised May, 20, 2022; Accepted June, 14, 2022.}
\thanks{This paper was recommended for publication by Editor Eric Marchand upon evaluation of the Associate Editor and Reviewers' comments.} 
\thanks{Part of this research was carried out at the Jet Propulsion Laboratory, California Institute of Technology, under a contract with the National Aeronautics and Space Administration. $\copyright$ 2021. All rights reserved.
The other part was carried out at the Robotics and Perception Group, University of Zurich, under contracts with the National Centre of Competence in Research (NCCR) Robotics through the Swiss National Science Foundation (SNSF) and the European Research Council (ERC) under grant agreement No. 51NF40\_185543.
We thank Konstantin Kalenberg for the feature prediction implementation improving EKLT's computational efficiency.}
\thanks{$^{1}$F. Mahlknecht, J. Nash, F. M. Rockenbauer, B. Morrell and J. Delaune are with the Jet Propulsion Laboratory, California Institute of Technology, USA.}%
\thanks{$^{2}$D. Gehrig and D. Scaramuzza are with the Robotics and Perception Group, University of Zurich, Switzerland {\tt\footnotesize https:/rpg.ifi.uzh.ch}}%
\thanks{Digital Object Identifier (DOI): see top of this page.}
}
%
%

\markboth{IEEE Robotics and Automation Letters. Preprint Version. Accepted June, 2022}
{Mahlknecht \MakeLowercase{\textit{et al.}}: Exploring Event Camera-based Odometry for Planetary Robots} 

%



\maketitle

\begin{abstract}
Due to their resilience to motion blur and high robustness in low-light and high dynamic range conditions, event cameras are poised to become enabling sensors for vision-based exploration on future Mars helicopter missions. However, existing event-based visual-inertial odometry (VIO) algorithms either suffer from high tracking errors or are brittle, since they cannot cope with significant depth uncertainties caused by an unforeseen loss of tracking or other effects. In this work, we introduce EKLT-VIO, which addresses both limitations by combining a state-of-the-art event-based frontend with a filter-based backend. This makes it both accurate and robust to uncertainties, outperforming event- and frame-based VIO algorithms on challenging benchmarks by 32\%. In addition, we demonstrate accurate performance in hover-like conditions (outperforming existing event-based methods) as well as high robustness in newly collected Mars-like and high-dynamic-range sequences, where existing frame-based methods fail. In doing so, we show that event-based VIO is the way forward for vision-based exploration on Mars.
\end{abstract}

\begin{IEEEkeywords}
Vision-Based Navigation; Space Robotics and Automation; Visual-Inertial SLAM
\end{IEEEkeywords}

\section*{Multimedia Material:} For code and dataset please visit \url{https://uzh-rpg.github.io/eklt-vio/}.

%
\IEEEpeerreviewmaketitle

\section{Introduction}
%
%
%
%
\begin{figure}[ht]
    \centering
    \begin{tabular}{cc}
    \multicolumn{2}{c}{\includegraphics[width=0.9\linewidth]{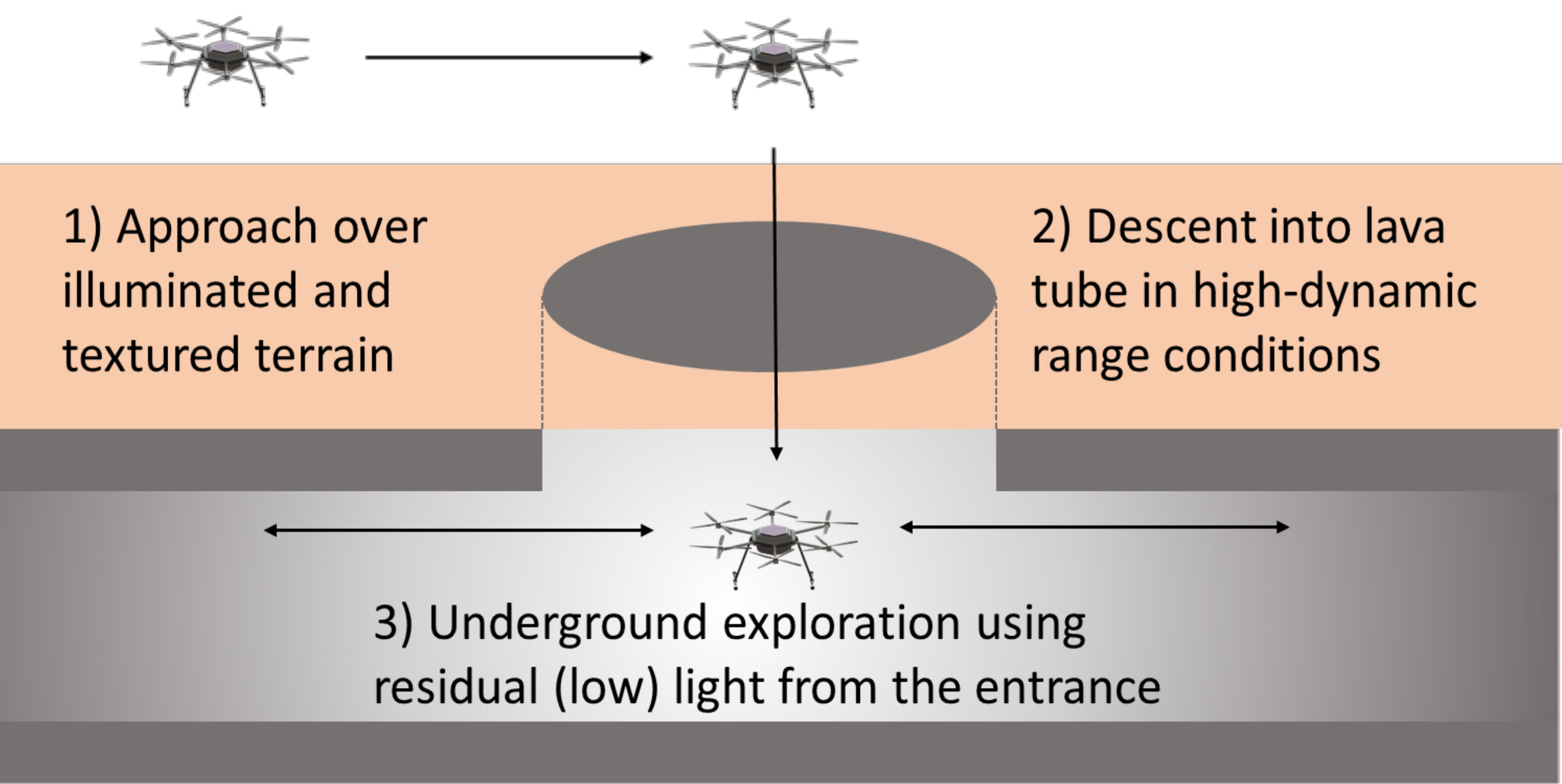}}\\
    \multicolumn{2}{c}{(a) Mission scenario}\\
    \includegraphics[height=0.25\linewidth]{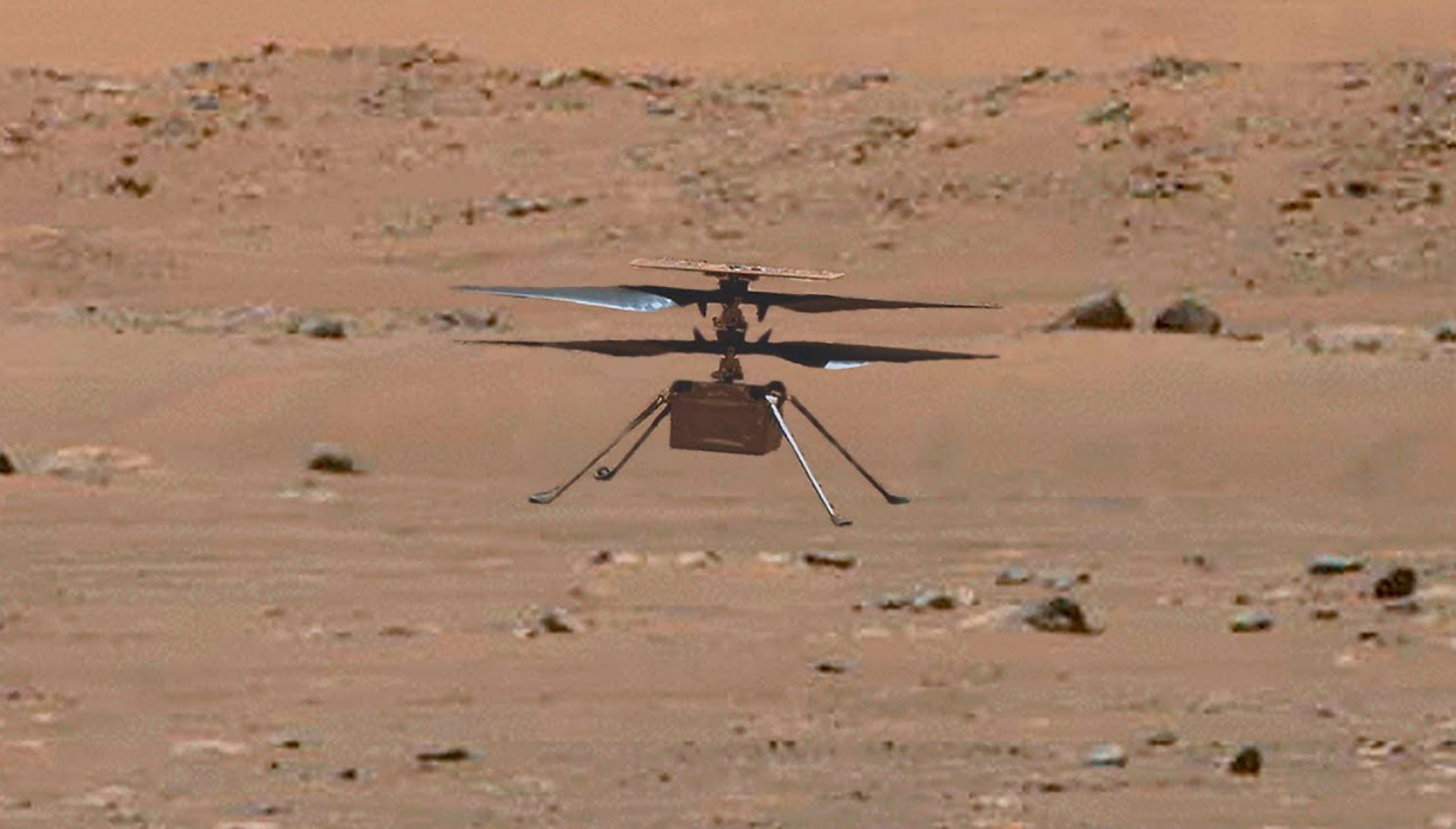} &
    \includegraphics[height=0.25\linewidth]{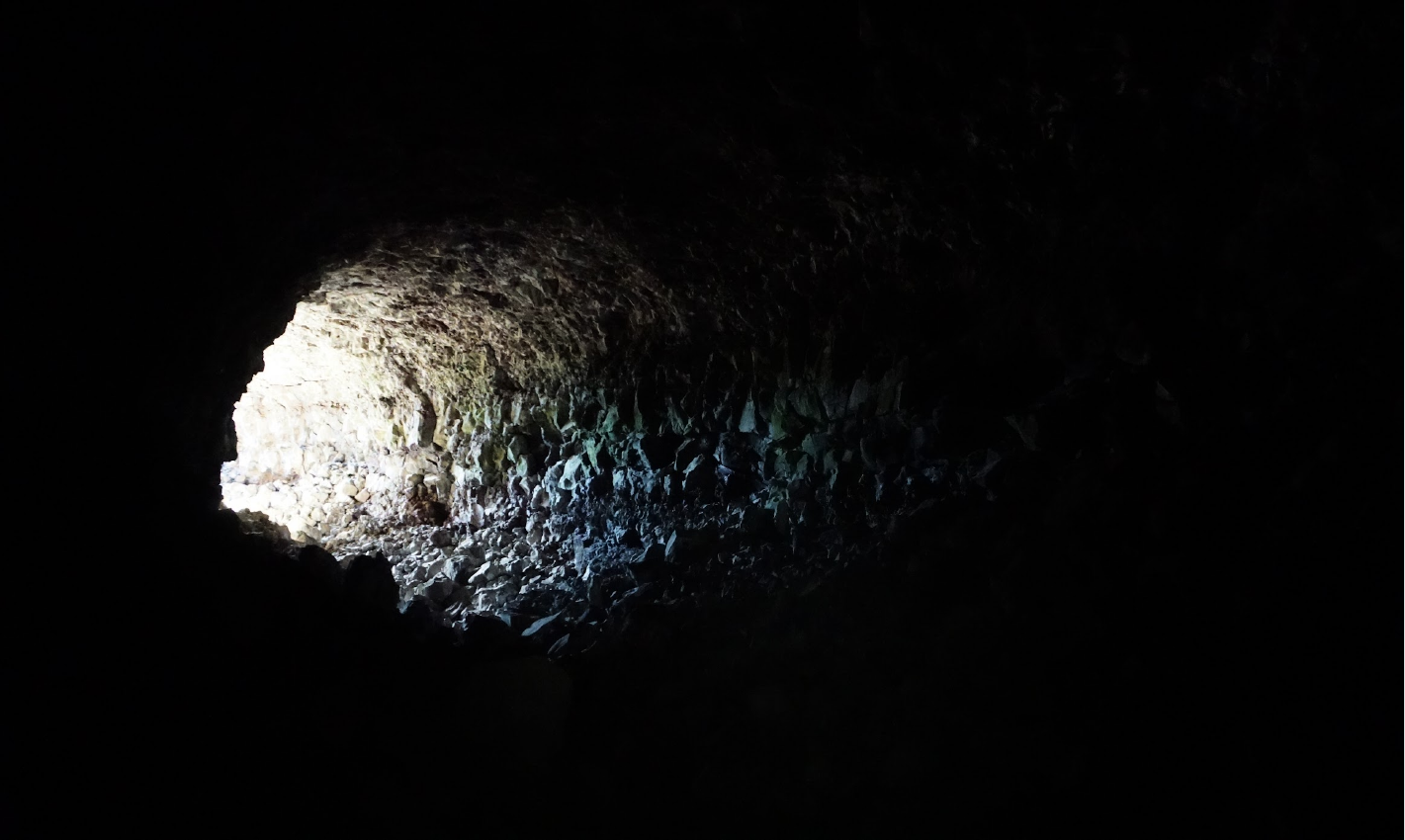}\\
    (b) Ingenuity Mars Helicopter & (c) Lava tube
    \end{tabular}
    \caption{New mission scenario (a) enabled by EKLT-VIO for a Mars helicopter (b) scouting the entrance of lava tubes (c).
    }
    \label{fig:conops}
    \vspace{-3ex}
\end{figure}

\IEEEPARstart{S}{tate} \rev{estimation is critical for enabling autonomous navigation and control of mobile robots, with widespread applications from space exploration to household cleaning robots. There exist well-established algorithms, such as~\cite{li2013msckfslam, Leutenegger15ijrr,Qin18tro,forster2016manifold} which estimate ego-motion from visual-inertial data.  However, vision-based navigation is drastically impacted by the known limitations of conventional cameras, such as motion blur and low dynamic range.}

Event cameras promise to address these limitations~\cite{gallego2019event}. Unlike a standard camera that measures absolute pixel brightness using a global exposure time, event camera pixels independently detect positive or negative brightness changes at microsecond resolution. Event cameras can provide data at 1 MHz and 120 dB dynamic range, both orders of magnitude greater than what can be achieved with a standard 60 dB camera. This leads to a significant reduction in motion blur, and enables operation in high dynamic range (HDR), low light, and fast motion conditions~\cite{Rosinol18ral,sun2021autonomous}.

On the application side, computer vision is increasingly used in modern planetary robotic missions~\cite{Johnson2017TheLV, bos2018osiris,bayard2019vision,Maimone07jfr,Delaune20ac}.
\revtwo{The resilient properties of event cameras may enable robots to explore in conditions where frame cameras cannot operate without introducing the size, weight, power, and range limitations of a 3D LiDAR}. 


In this paper, we focus on a scenario involving the exploration of the entrance of a lava tube by a Mars helicopter, as illustrated in Fig.~\ref{fig:conops}. Lava tubes are natural tunnels created by lava flows in volcanic terrains. Those found on Mars have drawn significant attention because of the possibility that they might host microbial life~\cite{carrier2020astro}. The natural protection from radiation offered by lava tubes also makes them candidates to host the first human base on Mars. 

Before sending a robotic mission~\cite{phillips2020macie} or astronauts to a specific lava tube, it would be desirable to scout and map several locations. 
Mars helicopters are candidate platforms to scout multiple lava tubes throughout a single mission. However, Mars helicopters cannot fly LiDARs and have to rely on passive cameras for navigation. Frame cameras are ill-suited to explore lava tubes because of the HDR conditions created by the shadow at the entrance of the tube, as well as the low-light conditions once inside. This capability gap is filled by event cameras, which offer the potential to explore and map the lava tube for potentially tens of meters using residual light from the entrance.
%

Mars helicopters come with their own requirements on the state estimation system~\cite{delaune2021xvio, bayard2019vision}. They must rely on small passive lightweight cameras to observe the full state up to scale and gravity direction. The camera is fused with an inertial measurement unit (IMU), which makes gravity observable, enables a high estimation rate, and acts as an emergency landing sensor in case of camera failure. Finally, a laser range finder is used to observe scale in the absence of accelerometer excitation. The estimation backend must be able to handle feature depth uncertainty associated with helicopter hovering and rotation-only dynamics. \revtwo{Due to this uncertainty successful feature triangulation is often inhibited in these cases, leading to failure of optimization-based backends, which critically rely on triangulated features. By contrast, filter-based approaches leverage priors to initialize depth measurements and thus do not suffer from this issue~\cite{delaune2020xvio}}. \rev{This proved critical in Ingenuity Mars helicopter's sixth flight on Mars, where an image timestamping anomaly caused roll and pitch oscillations greater than 20 degrees~\cite{grip2021}. Such rotations cause a loss of features, which can lead to estimation failure in non-filter-based state estimation approaches, which are fundamentally unable to handle the depth uncertainty of the new feature tracks without a dedicated re-initialization procedure.}

\rev{State-of-the-art event-based VIO methods are unsuitable in these conditions since they either \emph{(i)} use optimization-based backends, which do not model depth uncertainty, thus featuring brittle performance in mission-typical rotation-only motion, or when a significant portion of features are lost~\cite{Rosinol18ral}, or \emph{(ii)} show a higher tracking error, due to the use of suboptimal event-based frontends~\cite{Zhu17cvpr}.} \revthree{Image-based VIO methods such as \cite{delaune2021xvio,Kottas13iros} have addressed this by using depth priors~\cite{delaune2021xvio} or motion classification~\cite{Kottas13iros}.}

\rev{In this work, we introduce EKLT-VIO, which builds on the EKF backend in \cite{delaune2021xvio} which handles pure rotational motion, and combines it with the state-of-the-art event-based feature tracker EKLT~\cite{Gehrig19ijcv}, thereby addressing the limitations above.}
\rev{EKLT-VIO is accurate, outperforming previous state-of-the-art frame-based and event-based methods on the challenging Event-Camera Dataset~\cite{Mueggler17ijrr}, with a 32\% improvement in terms of pose accuracy.}
\revtwo{Moreover, by leveraging depth uncertainty it reduces its reliance on triangulating features, which both increase robustness during purely-rotational motion, and facilitates rapid initialization, both of which are limitations of  existing optimization-based methods.} This is because they require lengthy bootstrapping sequences, which would be impractical on Mars. Additionally, it maintains state-estimate, even when frame-based methods fail due to excessive motion blur.
We show that our event-based EKLT frontend has a higher tracking performance than existing methods on newly collected data in Mars-like conditions. This demonstrates the viability of our EKLT-VIO on Mars. Our contributions are:
\ifdefined\MYE
\fi
\begin{itemize}
    \item \rev{We introduce EKLT-VIO, an event-based VIO method that combines an accurate state-of-the-art event-based feature tracker EKLT} with an EKF backend. It outperforms state-of-the-art event- and frame-based methods, reducing the overall tracking error by 32\%.
    \item We show accurate and robust tracking even in rotation-only sequences, which are closest to the hover-like scenarios experienced by Mars helicopters, outperforming optimization-based and frame-based methods. 
    \item We outperform existing methods on newly collected Mars-like sequences collected in the JPL Mars Yard and Wells Cave for planetary exploration. 
\end{itemize}

\section{Related Work}

\textbf{Frame-based VIO:} An overview of existing approaches is discussed in \cite{Delmerico18icra}.
Frame-based VIO algorithms can be roughly segmented into two classes: optimization-based and filter-based algorithms~\cite{Delmerico18icra}. While both algorithms focus on tracking camera poses by minimizing both visual and inertial residuals, optimization-based methods solve this by performing iterative Gauss-Newton steps, while filtering-based methods achieve this through Kalman Filtering steps.  

Since optimizing both 3D landmarks (\emph{i.e.}, SLAM features) and camera poses is costly, several filtering-based techniques exist that focus on refining camera poses from bearing measurements (\emph{i.e.}, \rev{multi-state constraint Kalman filter (MSCKF) features \cite{Mourikis07icra}}) directly. However, MSCKF features need translational motion and provide updates only after the full feature track is known. The filtering-based approach, xVIO~\cite{delaune2021xvio}, combines the advantages of both features, with robustness to depth uncertainty in rotation-only motion and computational efficiency with many MSCKF features.\\

\begin{figure*}[t]
\vspace{4ex}
    \centering
    \includegraphics[width=0.8\linewidth]{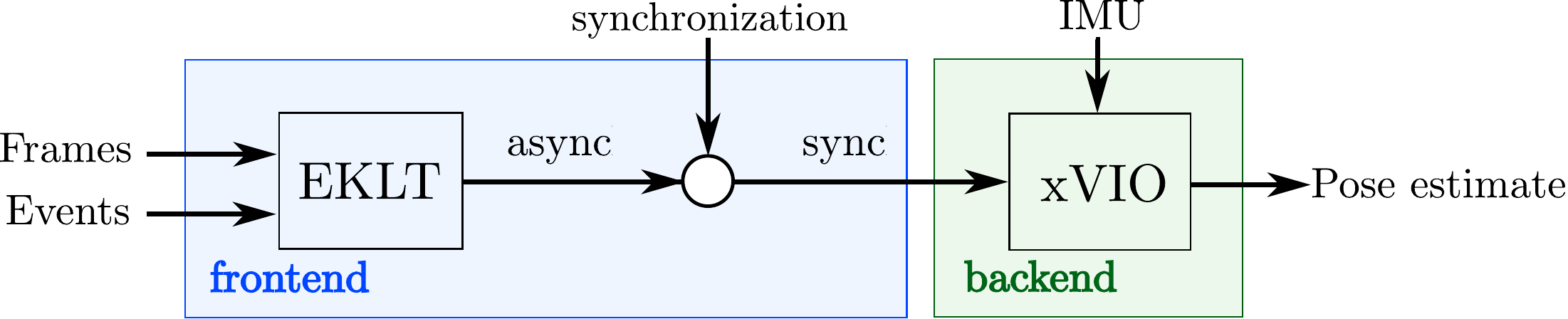}
    \caption{\rev{We combine the feature tracker EKLT, which use frames and events, with the filter-based backend xVIO to enable low-translation state-estimation. In contrast to standard, frame-based VIO, an additional synchronization step converts asynchronous tracks to synchronous matches, which are used by the backend. This enables variable-rate backend updates.}}
    \label{fig:overview}
    \vspace{-3ex}
\end{figure*}
\begin{figure}[ht]
    \centering
    \includegraphics[width=.8\linewidth]{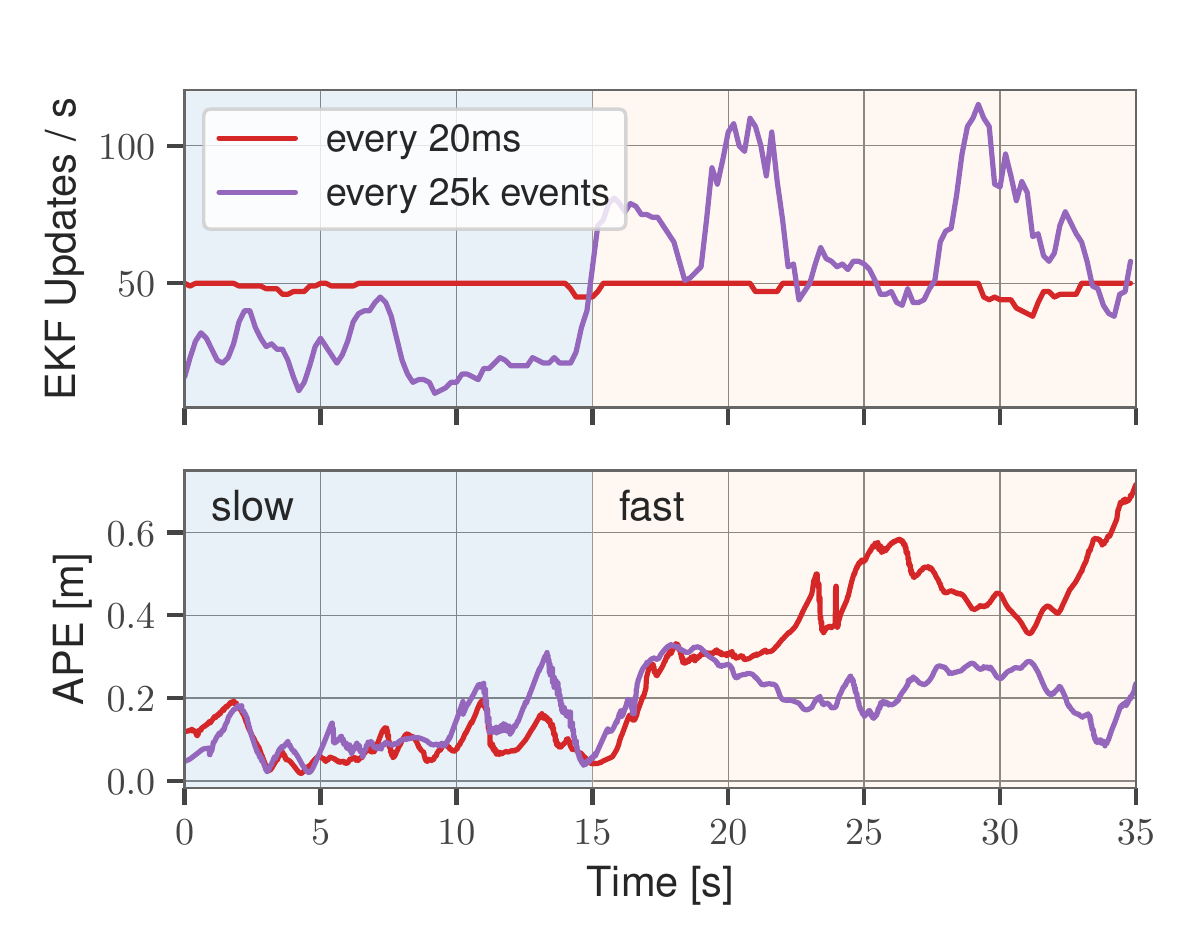}
    \caption{\rev{Synchronous feature updates (red) tend to generate too many updates during slow sequences and too few during fast sequences, leading to high tracking error. Our irregular update strategy (purple) adapts to the event-rate, and thus maintains low tracking error in both scenarios.}}
    \label{fig:update-strategy}
\end{figure}

\textbf{Event-based VIO:} First event-based, 6-DOF visual odometry (VO) algorithms only started to appear recently~\cite{Kim16eccv,Rebecq17ral}. 
Later work incorporated an IMU to improve tracking performance and stability~\cite{Rebecq17bmvc,Zhu17cvpr}, achieving impressive tracking on a fast spinning leash~\cite{Rebecq17bmvc}.
Despite their robustness, these methods are affected by drift due to the differential nature of the used sensors. This is why Ultimate SLAM (USLAM)~\cite{Rosinol18ral} used a combination of events, frames, and IMU, all provided by the Dynamic and Active Vision Sensor (DAVIS)~\cite{Brandli14ssc}. 
It tracks FAST corners~\cite{Rosten06eccv} on frames and motion-compensated event frames separately using the Lucas-Kanade tracker (KLT)~\cite{lucas1981iterative} and fuses these feature tracks with IMU measurements in a sliding window.

While addressing drift, USLAM still relies on a sliding window optimization scheme, which is expensive and does not allow pose-only optimization through the use of MSCKF features.
Moreover, its FAST/KLT frontend, first introduced in~\cite{Rebecq17bmvc}, is optimized explicitly for frame-like inputs and was shown to transfer suboptimally to event-based frames~\cite{Gehrig19ijcv}.
\rev{In this work, we incorporate the state-of-the-art event-based tracker EKLT~\cite{Gehrig19ijcv}, which takes a more principled approach to fusing events and frames, and thus achieves better feature tracking performance compared to \cite{Rosinol18ral,Rebecq17bmvc}}.

\section{Methodology}
\rev{In this section we present EKLT-VIO, which is illustrated in Fig.~\ref{fig:overview}.
It is an event-based VIO algorithm based on the state-of-the-art event tracker EKLT, coupled with a filter-based xVIO backend.}

\subsection{Backend}
\label{sub:backend}
We start by providing a summary of the xVIO backend.
For more details see~\cite{delaune2021xvio}. The backend fuses data from an inertial measurement unit (IMU) and feature tracks from the frontend.
It does this by using an extended Kalman filter (EKF) with an IMU state $\mathbf{x}_I$ and a visual state $\mathbf{x}_V$:

\begin{equation}
    \mathbf{x} = \begin{bmatrix} {\mathbf{x}_I}^\intercal & {\mathbf{x}_V}^\intercal \end{bmatrix}^\intercal
\end{equation}

The IMU state follows an inertial propagation scheme as described in~\cite{Weiss12icraRealtime}. The visual state $\mathbf{x}_V$ is split into sliding window states $\mathbf{x}_S$ and feature states $\mathbf{x}_F$:

\begin{align}
    \mathbf{x}_V &= \begin{bmatrix} {\mathbf{x}_F}^\intercal & {\mathbf{x}_S}^\intercal \end{bmatrix}^\intercal,\quad \mathbf{x}_F = \begin{bmatrix} \mathbf{f}_1 & \hdots & \mathbf{f}_N  \end{bmatrix}^\intercal\\
    \mathbf{x}_S &= \begin{bmatrix} {\mathbf{p}_w^{c_1}}^\intercal & \hdots & {\mathbf{p}_w^{c_M}}^\intercal & {\mathbf{q}_w^{c_1}}^\intercal & \hdots & {\mathbf{q}_w^{c_M}}^\intercal \end{bmatrix}^\intercal 
\end{align}

The sliding window states contain the positions, $\mathbf{p}_w^{c_i}$, and attitudes parameterized as quaternions, $\mathbf{q}_w^{c_i}$, of the last $M$ camera poses $\{c_i\}$ with respect to a world frame $\{w\}$.
The feature states contain the 3D positions, $\mathbf{f}_j$, of $N$ SLAM features. In this work $N=15$ and $M=10$.

\revthree{We use a discrete-time VIO approach, as opposed to one based on splines~\cite{Mueggler18tro,CioffiRal2022,Li14ijrrb,Eckenhoff21tro}. Although they can incorporate event-data~\cite{Mueggler18tro} more elegantly they are notoriously computationally expensive~\cite{Mueggler18tro} and less established. This is why we opt for discrete-time VIO and leave splines for future work.} 

SLAM features are parametrized with respect to an anchor pose $\mathbf{p}_w^{c_{a_j}}$ in the sliding window, and defined as $\mathbf{f}_j~=~[\alpha_j\quad \beta_j\quad \rho_j]$  with $\alpha_j$ and $\beta_j$ being normalized image coordinates and $\rho_j$ being the inverse depth. 
Each time the feature tracks are updated, each SLAM feature $j$ is converted from inverse-depth to Cartesian coordinates in the associated anchor camera frame $\{c_{a_j}\}$.

\begin{equation}
    \mathbf{p}_{c_{i}}^j = \mathbf{C}(\mathbf{q}_w^{c_{i}})
    \left(\mathbf{p}_w^{c_{a_j}} + \frac{1}{\rho_j} \mathbf{C}(\mathbf{q}_w^{c_{a_j}})^\intercal
    \begin{bmatrix}
     \alpha_j \\ \beta_j \\ 1
    \end{bmatrix}
    - \mathbf{p}_w^{c_i}
    \right),  \label{eq:slam-coordinate-transform}
\end{equation}


The measurement model is the normalized feature:


\begin{equation}
    \mathbf{z}_j = \pi(\mathbf{p}_{c_{i}}^j) + \mathbf{n}_j,\quad \pi(\mathbf{x})= \begin{bmatrix}
     x_1/x_3 & x_2/x_3
    \end{bmatrix}^T, \label{eq:feature-measurement}
\end{equation}

where $\pi(\mathbf{x})$ performs feature projection, $\mathbf{n}_j$ is Gaussian noise, and $\mathbf{z}_j$
are the new feature observations by the frontend, expressed in normalized image coordinates.
Eqs.~\eqref{eq:slam-coordinate-transform} and~\eqref{eq:feature-measurement} can be used to develop the EKF update by linearizing the SLAM feature reprojection.
Details are given in~\cite{delaune2021xvio}.

In addition to SLAM features, the backend maintains MSCKF features that additionally constrain the camera poses without an explicit inverse depth. 
MSCKF features are thus not part of the state, resulting in a smaller computational cost per feature.
They need to be observed for the last $2 \leq m \leq M$ frames, providing a corresponding observation for each pose in the sliding window.
MSCKF features require triangulation using those pose priors, so they can only be processed once a track with significant translation is observed.
\revtwo{Successfully triangulated MSCKF features are used to initialize SLAM features. When there is insufficient translation for triangulation, xVIO instead initializes the inverse depth with $\rho_0=\frac{1}{2d_\text{min}}$ and uncertainty $\sigma_0=\frac{1}{4d_\text{min}}$, corresponding to a semi-infinite depth prior, and discards the MSCKF feature track~\cite{Montiel06rss}. This depth prior is especially useful during pure rotation or initialization, where few features can be triangulated, since it can directly contribute to reducing the state covariance}.  
\subsection{Frontend}

Here we provide a summary of our EKLT frontend, and refer the reader to~\cite{Gehrig19ijcv} for more details. EKLT tracks Harris corners, extracted on frames, by aligning the predicted and measured brightness increment in a patch around the corners. It minimizes the normalized distance between these patches to recover the warping parameters $\mathbf{p}$ and normalize optical flow $\mathbf{v}$ as

\begin{equation}
    \{\mathbf{p}, \mathbf{v}\}=\argmin_{\mathbf{p}, \mathbf{v}}
    \norm{\normed{\Delta L (\mathbf{u})} - \normed{\Delta \hat{L}(\mathbf{u}, \mathbf{p}, \mathbf{v})}}. \label{eq:eklt-optimization}
\end{equation}

While $\Delta L$ is defined as an aggregation of events in a local patch, $\Delta\hat L$ is defined as the negative dot product between the local log image gradient and optical flow vector, following the linearized event generation model~\cite{Gallego17pami}.
Here $W(\mathbf{u}, \mathbf{p})$ aligns the image gradient with the measured brightness increments according to the alignment parameters $p$. 
\revtwo{EKLT minimizes Eq.~\eqref{eq:eklt-optimization} using Gauss-Newton and the Ceres library\cite{ceres-solver}, and recovers alignment parameters $p$ and optical flow $v$}. As opposed to the reference implementation of EKLT, which optimizes in a sliding window fashion after a fixed number of events, we trigger the optimization only when the adaptive number of events is reached, using each event batch only once.
This entails a significant speed-up without loss in accuracy.




\subsection{Frontend Adaptations}
\label{sub:frontend_adaptations}
\textbf{Asynchronous feature updates}: We convert the asynchronous feature tracks provided by EKLT to synchronous feature tracks via a synchronization step (Fig. \ref{fig:overview}. This step produces a temporally synchronized list of feature positions, which are passed to the backend. 
The backend uses the associated correspondences $\textbf{z}_i\Longleftrightarrow \textbf{z}_j$ together with consecutive camera poses $c_{i}$ and $c_{j}$ to update the state as discussed in Sec.~\ref{sub:backend}. 
It is performed by selecting the most recent feature in the currently tracked feature set and extrapolating the positions of all other features to its timestamp. We synchronize every time, a fixed number of events $n_e$ is triggered, enabling variable-rate backend updates.
\revthree{We empirically found $n_e=3200$ to work best, see Tab. \ref{tab:update-thresholds}. We argue that reducing $n_e$ will introduce additional noisy updates to the EKF which reduce the accuracy, while having too high $n_e$ makes our approach less robust during high-speed motion.}

\begin{table}[H]
\centering
\begin{tabular}{@{}r|rrrrrrrr@{}}
\toprule
$n_e$ & 500 & 1000 & 3200 & 4800 & 7200 & 9200 & 15000 & 20000 \\
MMPE & 0.57 & 0.55 & \textbf{0.49} & 0.59 & 0.60 & 0.68 & 0.83 & 1.72 \\ \bottomrule
\end{tabular}
\caption{\revthree{Median Mean Position Error (MMPE) {[}\%{]} on the Event Camera Dataset for different EKF event update thresholds}}
\label{tab:update-thresholds}
\end{table}

\rev{This variable rate allows our algorithm to adapt to the scene dynamics (Fig. \ref{fig:update-strategy}), leading to fewer EKF updates in slow sequences (Fig.\ref{fig:update-strategy}, left) and a lower tracking error during high-speed sequences, compared to fixed rate updating. These features motivate the use of an event-based frontend since a purely frame-based one is limited by the framerate of the camera.} \revthree{Although, this may lead to drift in purely stationary environments where no events are triggered, this can easily be amended by enforcing a minimal backend update rate. Or by enforcing a no-motion prior when the event rate goes below a threshold, as in \cite{Rosinol18ral}.} 

\textbf{Outlier rejection}: For EKLT we exclusively reject outliers by setting a maximum threshold on the optimized residual of the alignment score in Eq.~\eqref{eq:eklt-optimization}. This allows outliers to be rejected quickly, without the need for costly geometric verification, such as 8-point RANSAC.

\section{Experiments}
\rev{
We start by validating our approach on standard benchmarks in Sec. \ref{sub:real}, where we compare the performance of EKLT-VIO against state-of-the-art event-based~\cite{Zhu17cvpr}, frame-based~\cite{delaune2021xvio} and event- and frame-based methods~\cite{Rosinol18ral}. To study the effect on the event-based feature tracker, we also study an additional baseline, based on the HASTE feature tracker~\cite{alzugaray:3DV19}. 
We then proceed to demonstrate the suitability of our approach on two important use-cases motivated by the Mars exploration scenario: \emph{(i)} pure rotational motion, imitating hover-like conditions on Mars (Sec. \ref{sub:rotation}), and \emph{(ii)} challenging HDR conditions on newly collected datasets in the JPL Mars Yard and at the entrance of the Wells Cave, emulating the entry into lava tubes (Sec. \ref{sub:cavern}). 
}
\rev{
\subsection{Baselines and Compared Methods}
\noindent\textbf{USLAM}~\cite{Rosinol18ral} is an event- and frame-based VIO method, which fuses feature tracks derived from frames and event-frames in an optimization-based backend.\\
\textbf{EVIO}~\cite{Zhu17cvpr} uses only events and IMU. Events are used to generate asynchronous feature tracks, which are then fused in a filter-based backend. Since open-source code is not available, we only report results on real sequences.\\
\textbf{KLT-VIO}~\cite{delaune2021xvio} is a frame-based VIO method that fuses feature tracks based on FAST/KLT in a filter-based backend, and is specifically designed for use during helicopter flight.\\
\textbf{HASTE-VIO}~\cite{alzugaray:3DV19} Finally, we combine the state-of-the-art purely event-based tracker HASTE~\cite{alzugaray:3DV19} with xVIO as an additional baseline.
Similar to EKLT, it produces asynchronous feature tracks which are first synchronized using the method described in Sec. \ref{sub:frontend_adaptations}, before being fed into the backend.
}
\begin{table*}[t]
\vspace{4ex}
\centering
\resizebox{0.9\textwidth}{!}{%
\begin{tabular}{@{}lcccccccccccc@{}}
\toprule
Dataset &\multicolumn{2}{c}{\lightgray{USLAM*~\cite{Rosinol18ral}}} &  \multicolumn{2}{c}{USLAM~\cite{Rosinol18ral}} &\multicolumn{2}{c}{EVIO~\cite{Zhu17cvpr}} & \multicolumn{2}{c}{KLT-VIO~\cite{delaune2021xvio}} & \multicolumn{2}{c}{HASTE-VIO} & \multicolumn{2}{c}{EKLT-VIO (\textbf{ours})} \\ \midrule
 & \lightgray{MPE} & \lightgray{MYE} & MPE & MYE & MPE & MYE & MPE  & MYE & MPE  & MYE & MPE  & MYE\\ \midrule
Boxes 6DOF          &\lightgray{0.30}&\lightgray{0.04}& \textbf{0.68} & \textbf{0.03} &\rev{4.13} & \rev{0.92} & 0.97 & 0.05 & 2.03 & \multicolumn{1}{c|}{\textbf{0.03}} & 0.84 & 0.09  \\
Boxes Translation   &\lightgray{0.27}&\lightgray{0.02}&  1.12 & 2.62 &\rev{3.18} & \rev{0.67} & \textbf{0.33} & \textbf{0.08} & 2.55 & \multicolumn{1}{c|}{0.46} & 0.48 & 0.25  \\
Dynamic 6DOF        &\lightgray{0.19}&\lightgray{0.10}&  0.76 & 0.09 &\rev{3.38} & \rev{1.20} & 0.78 & \textbf{0.03} & \textbf{0.52} & \multicolumn{1}{c|}{0.06} & 0.79 & 0.06  \\
Dynamic Translation &\lightgray{0.18}&\lightgray{0.15}&  0.63 & 0.22 &\rev{1.06} & \rev{0.25} & 0.55 & 0.06 & 1.32 & \multicolumn{1}{c|}{0.06} & \textbf{0.40} & \textbf{0.04} \\
HDR Boxes           &\lightgray{0.37}&\lightgray{0.03}&  1.01 & 0.31 &\rev{3.22} & \rev{0.15} & \textbf{0.42} & \textbf{0.02}  & 1.75 & \multicolumn{1}{c|}{0.09} & 0.46 & 0.06 \\
HDR Poster          &\lightgray{0.31}&\lightgray{0.05}&  1.48 & 0.09 &\rev{1.41} & \rev{0.13}& 0.77 & 0.03  & 0.57 & \multicolumn{1}{c|}{\textbf{0.02}} & \textbf{0.65} & 0.04 \\
Poster 6DOF         &\lightgray{0.28}&\lightgray{0.07}&  0.59 & 0.03 &\rev{5.79} & \rev{1.84} & 0.69 & \textbf{0.02}  & 1.50 & \multicolumn{1}{c|}{0.03} & \textbf{0.35} & \textbf{0.02} \\
Poster Translation  &\lightgray{0.12}&\lightgray{0.04}&  0.24 & \textbf{0.02} &\rev{1.59} & \rev{0.38} & \textbf{0.16} & \textbf{0.02} & 1.34 & \multicolumn{1}{c|}{\textbf{0.02}} & 0.35 & 0.03 \\
Shapes 6DOF         &\lightgray{0.10}&\lightgray{0.04}&  1.07 & 0.03 &\rev{2.52} & \rev{0.61} & 1.80 & 0.03 & 2.35 & \multicolumn{1}{c|}{\textbf{0.02}} & \textbf{0.60} & 0.03 \\
Shapes Translation  &\lightgray{0.26}&\lightgray{0.06}& 1.36 & \textbf{0.01} &\rev{4.56} & \rev{2.60} &  1.38 & 0.02 & 1.09 & \multicolumn{1}{c|}{0.02} & \textbf{0.51} & 0.03 \\ \bottomrule
\textbf{Average}    &\lightgray{0.24}&\lightgray{0.06}& 0.89 & 0.34 &\rev{3.08} & \rev{0.88} & 0.79 & \textbf{0.04} & 1.50 & \multicolumn{1}{c|}{0.08} & \textbf{0.54} & 0.07\\
\multicolumn{13}{l}{\rev{\noindent*per-sequence hyperparameter tuning and correct IMU bias intialization}}\\\bottomrule
\end{tabular}}
\caption{Pose estimate accuracy comparison on the Event-Camera Dataset~\cite{Mueggler17ijrr} in terms of mean position error (MPE) in \% and mean yaw error (MYE) in deg/m. \rev{Grayed-out results with (*)} by USLAM~\cite{Rosinol18ral} were achieved through per-sequence parameter tuning and correct IMU bias initialization, while results in black used a single parameter set, tuned on all sequences simultaneously, and were initialized with an IMU bias of zero. 
}
\label{tab:rpg-davis-data}
\end{table*}
\begin{table*}[t]
\centering
\begin{tabular}{@{}lcccccccc@{}}
\toprule
Dataset & \multicolumn{2}{c}{USLAM~\cite{Rosinol18ral}} &\multicolumn{2}{c}{KLT-VIO~\cite{delaune2021xvio}} & \multicolumn{2}{c}{HASTE-VIO}  & \multicolumn{2}{c}{EKLT-VIO (\textbf{ours})}\\ \midrule
& MPE & MYE & MPE & MYE & MPE & MYE & MPE & MYE \\ \midrule
Dynamic Rotation & \multicolumn{2}{c}{\multirow{4}{*}{\textit{unfeasible}}}                                                                                                                                        & 9.97                                                                        & \textbf{0.13}                                                                   & \textbf{6.22}                                                               & \multicolumn{1}{c|}{2.32}& 7.71                                                                        & 1.52                                                                                                                                                        \\
Boxes Rotation   & \multicolumn{2}{c}{}                                                                                                    & \multicolumn{2}{c}{\textit{diverging}}                                                                                                                        & 20.57                                                                       & \multicolumn{1}{c|}{\textbf{1.32}} & \textbf{8.78}                                                               & 1.36                                                                                                                                               \\
Poster Rotation  & \multicolumn{2}{c}{}                                                                                                                                        & \multicolumn{2}{c}{\textit{diverging}}                                                                                                                        & 3.96                                                                        & \multicolumn{1}{c|}{\textbf{0.09}}                                                                   & \textbf{1.44}                                                               & \textbf{0.09}                                                                   \\ 
Shapes Rotation  & \multicolumn{2}{c}{}                                                                                                                                        & \multicolumn{2}{c}{\textit{diverging}}                                                                                                                        & \multicolumn{2}{c|}{\textit{diverging}}& \textbf{6.95}                                                               & \textbf{4.59}                                                                                                                                                                                           \\\bottomrule
\end{tabular}%
\caption{Mean position and yaw error (MPE and MYE) in \% and deg/m on rotation-only sequences.}
\label{tab:rpg-davis-rotation}
\vspace{-3ex}
\end{table*}
\subsection{Real Data}
\label{sub:real}
We benchmark our methods on the Event-Camera Dataset~\cite{Mueggler17ijrr}
, recorded with a DAVIS 240C~\cite{Brandli14ssc} with synchronized images, events, IMU measurements, and very fast hand-held motions in an HDR scenario. An OptiTrack is used for ground-truth camera trajectories.
We evaluate the pose tracking accuracy using the same protocol as~\cite{Rosinol18ral}, and report mean position error (MPE) in \% of the total trajectory length and mean yaw error (MYE) in deg/m in Tab.\ref{tab:rpg-davis-data}.

In~\cite{Rosinol18ral}, USLAM uses different parameters for each sequence, and correct IMU bias initialization, resulting in the gray columns in Tab.~\ref{tab:rpg-davis-data}. \rev{We mark this method as USLAM*}. However, on Mars, VIO systems should perform robustly in unknown environments, making, parameter tuning and bias initialization infeasible. For this reason, we retune the parameters of USLAM to perform best on all sequences simultaneously resulting in the black values in Tab.~\ref{tab:rpg-davis-data}. All other methods were tuned in the same way.
\rev{Comparing USLAM* with USLAM shows that IMU bias initialization, and per-sequence hyperparameter tuning are clearly important to achieve low tracking error, reducing the error from 0.89\% to 0.24\%. Our EKLT-VIO, on the other hand, achieves an average error of 0.54\% without bias initialization, 39\% lower than USLAM. This improvement indicates that EKLT-VIO is simultaneously more robust to zero IMU bias initialization, and per-sequence hyperparameter tuning. }
In terms of position error, EKLT-VIO outperforms all other methods on 5 out of 10 sequences. With an average MPE of 0.54\% EKLT-VIO shows a 32\% lower MPE than runner-up KLT-VIO with 0.79\%. 
\rev{Finally, with a 3.08\% MPE, EVIO~\cite{Zhu17cvpr} is outperformed by EKLT-VIO by 82\%. 
} 

\subsection{Rotation-only sequences}
\label{sub:rotation}
\rev{As a next step, we show the suitability of EKLT-VIO in a Mars Mission-like scenario.  
To do this, we evaluate all methods on the rotation-only sequences of the Event-Camera Dataset, which are challenging for optimization-based backends such as USLAM~\cite{Rosinol18ral}. Similar to hover-like conditions expected during Mars missions, these sequences translate only little compared to the average scene depth, which poses a challenge for keyframe generation and triangulation. 

\begin{figure*}[h]
\centering
\vspace{4ex}
\begingroup
\setlength{\tabcolsep}{4pt} 
\renewcommand{\arraystretch}{1} 
\begin{tabular}{cccc}
  \includegraphics[trim=504 0 504 0, clip,height=.20\linewidth]{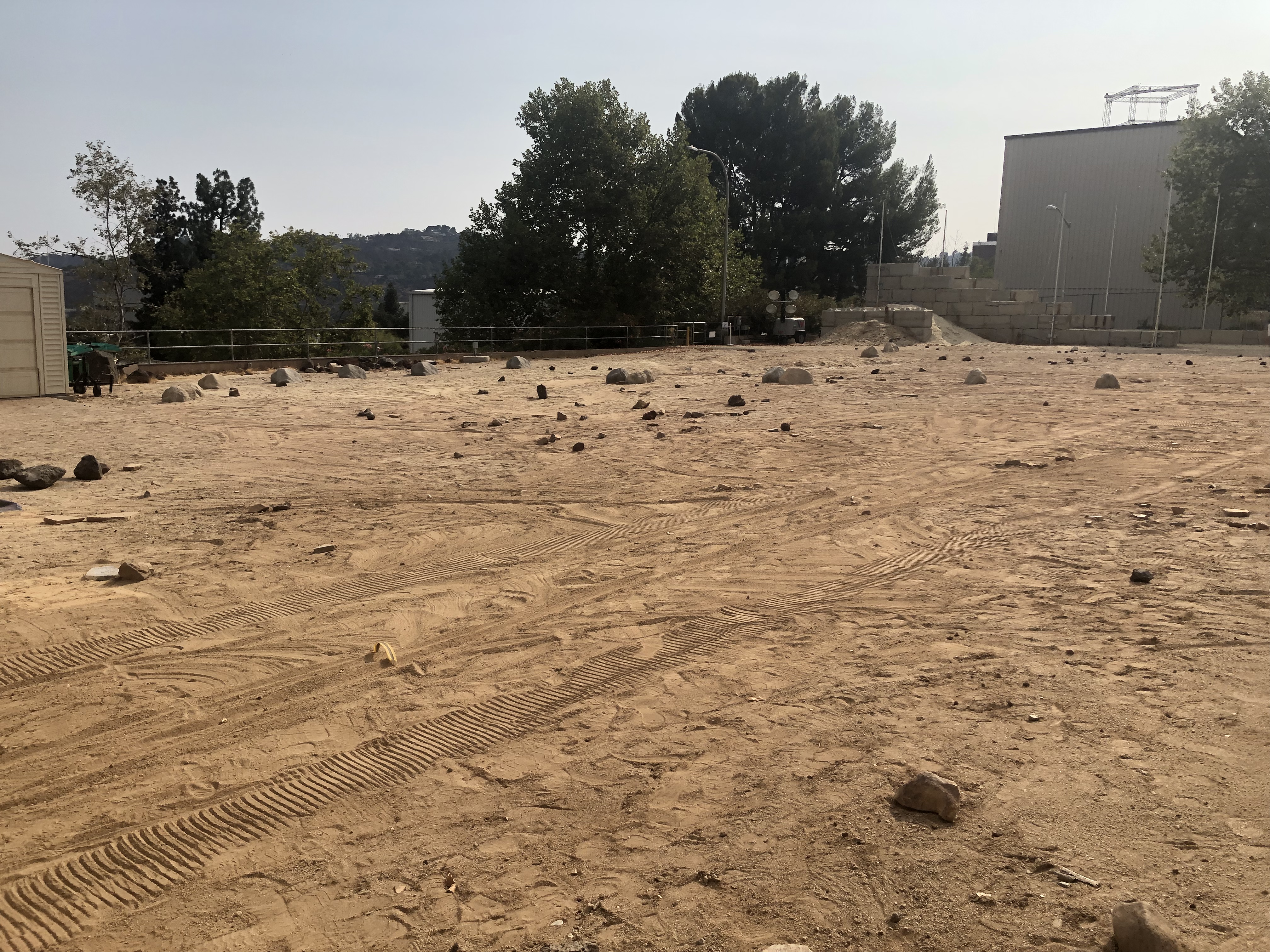}&
  \includegraphics[trim=286 0 286 0, clip,height=.20\linewidth]{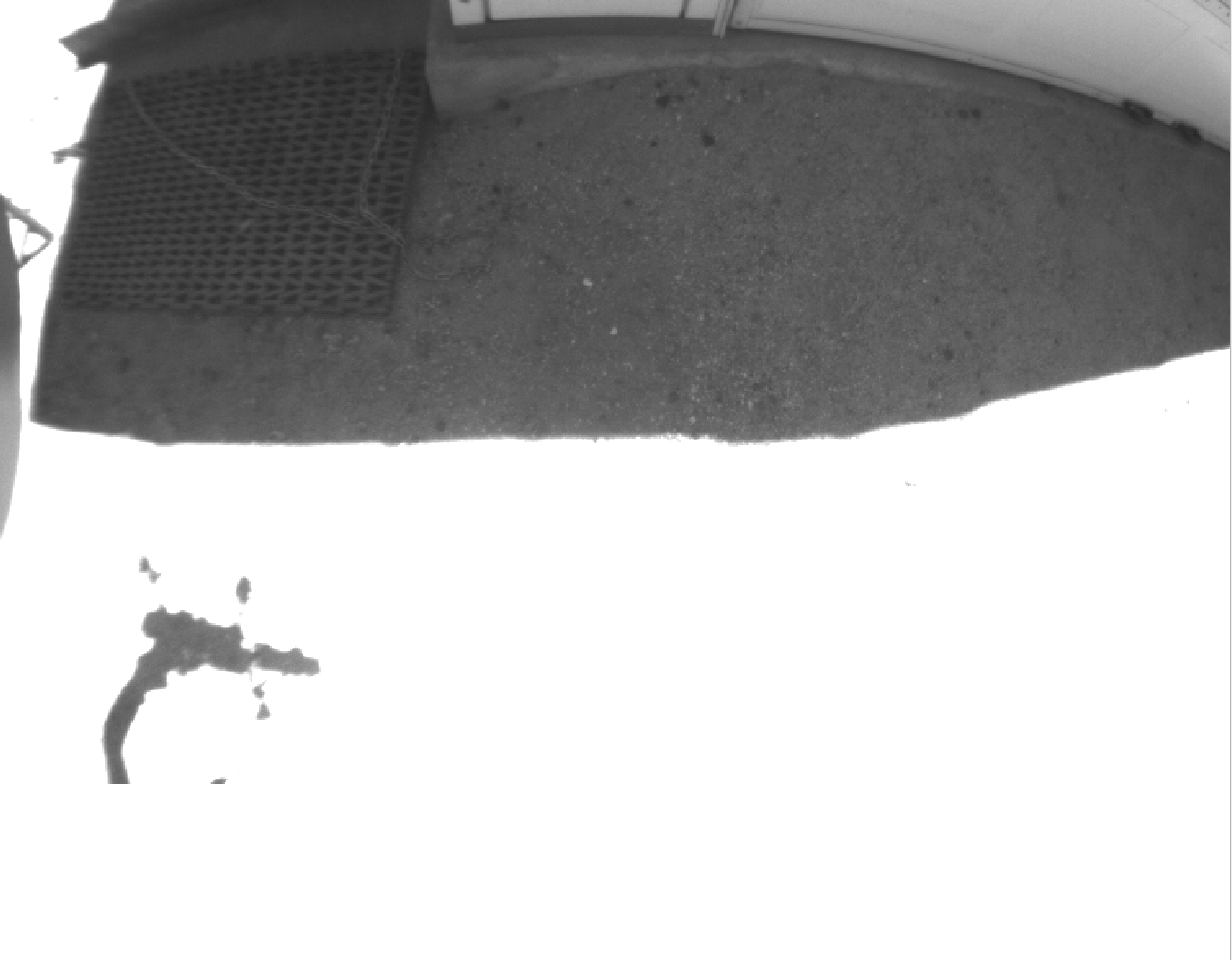}&
  \includegraphics[trim=80 0 80 0, clip,height=.20\linewidth]{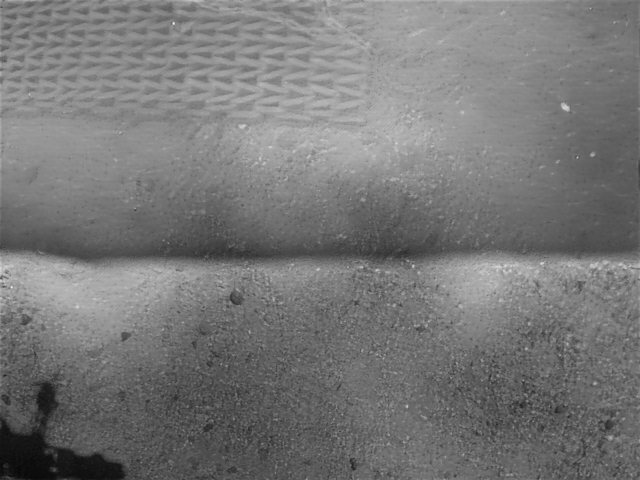}&
  \includegraphics[height=.20\linewidth]{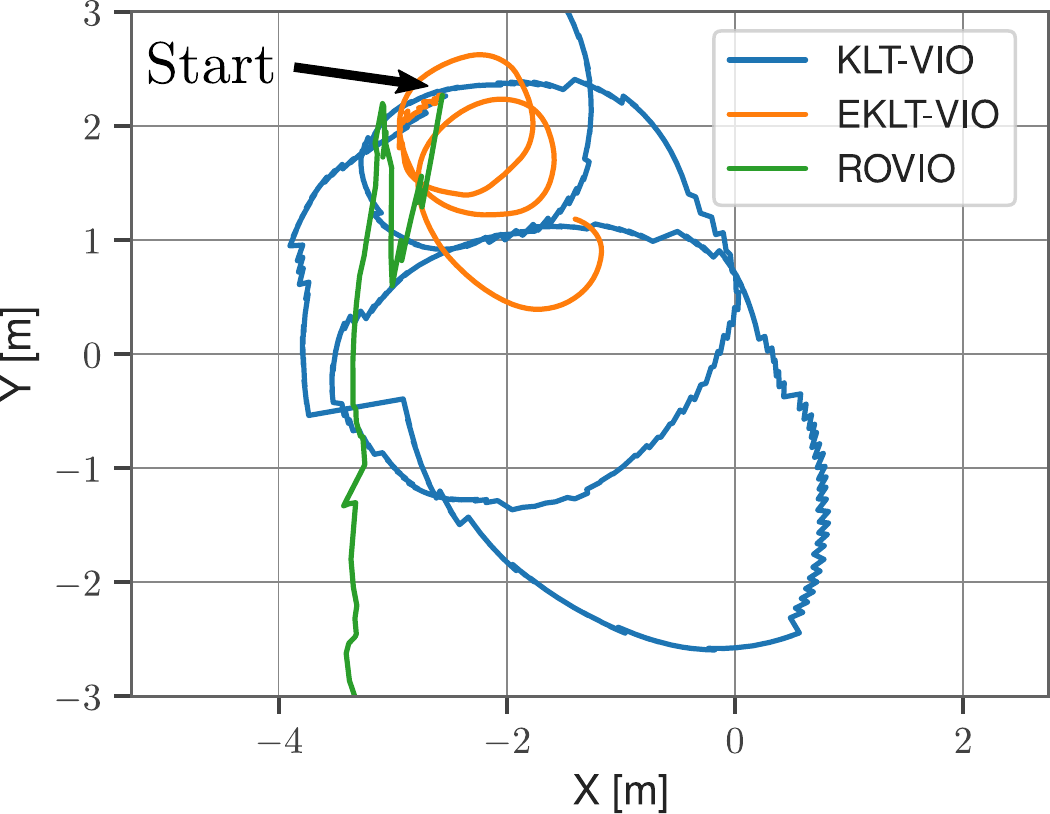}\\
  (a) Mars Yard preview & (b) Overexposed image & (c) Recons. from events & (d) Trajectories\\
  \includegraphics[trim=64 0 64 0, clip,height=.20\linewidth]{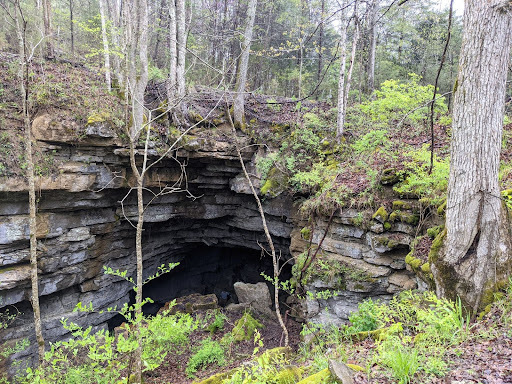}&
  \includegraphics[trim=23 0 23 0, clip,height=.20\linewidth]{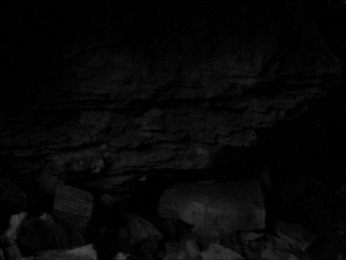}&
  \includegraphics[trim=25 0 25 0, clip,height=.20\linewidth]{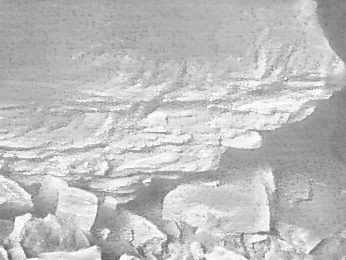}&
  \includegraphics[height=.20\linewidth]{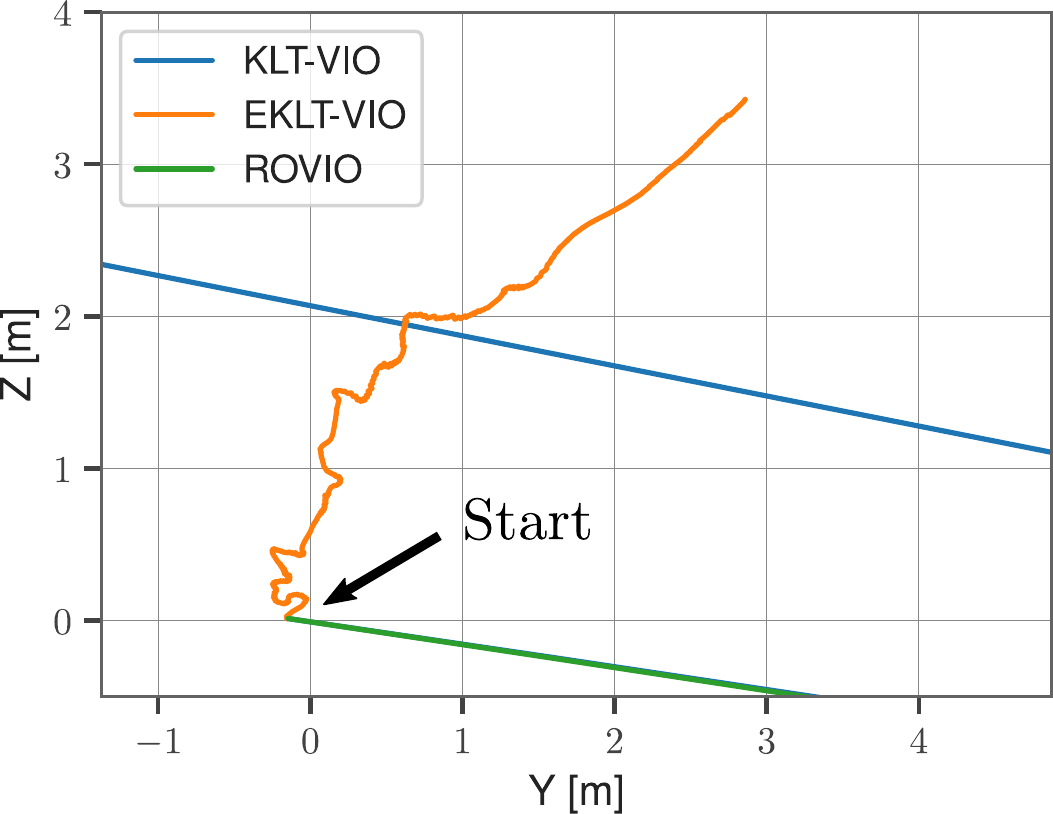}\\
    (e) Wells Cave preview & (f) Underexposed image & (g) Recons. from events & (h) Trajectories
  
\end{tabular}
\endgroup
\caption{\rev{In the Mars Yard (a) we test HDR conditions which cause severe oversaturation artefacts in standard images (b). Instead in the Wells Cave (e) we study low light scenarios encountered in lava tubes, which cause undersaturation (f). HDR images reconstructed from events~\cite{Rebecq19pami} (c,g) do not suffer from these artefacts, and are used by our method. As a result, we outperform existing frame-based approaches KLT-VIO~\cite{delaune2021xvio} and ROVIO~
\cite{Bloesch15iros} on both trajectories. }} 
\label{fig:my_comp}
\vspace{-2ex}
\end{figure*}

\begin{figure}
\centering
\begingroup
\setlength{\tabcolsep}{2pt} 
\renewcommand{\arraystretch}{.5} 
\begin{tabular}{c}
     \includegraphics[width=0.8\linewidth]{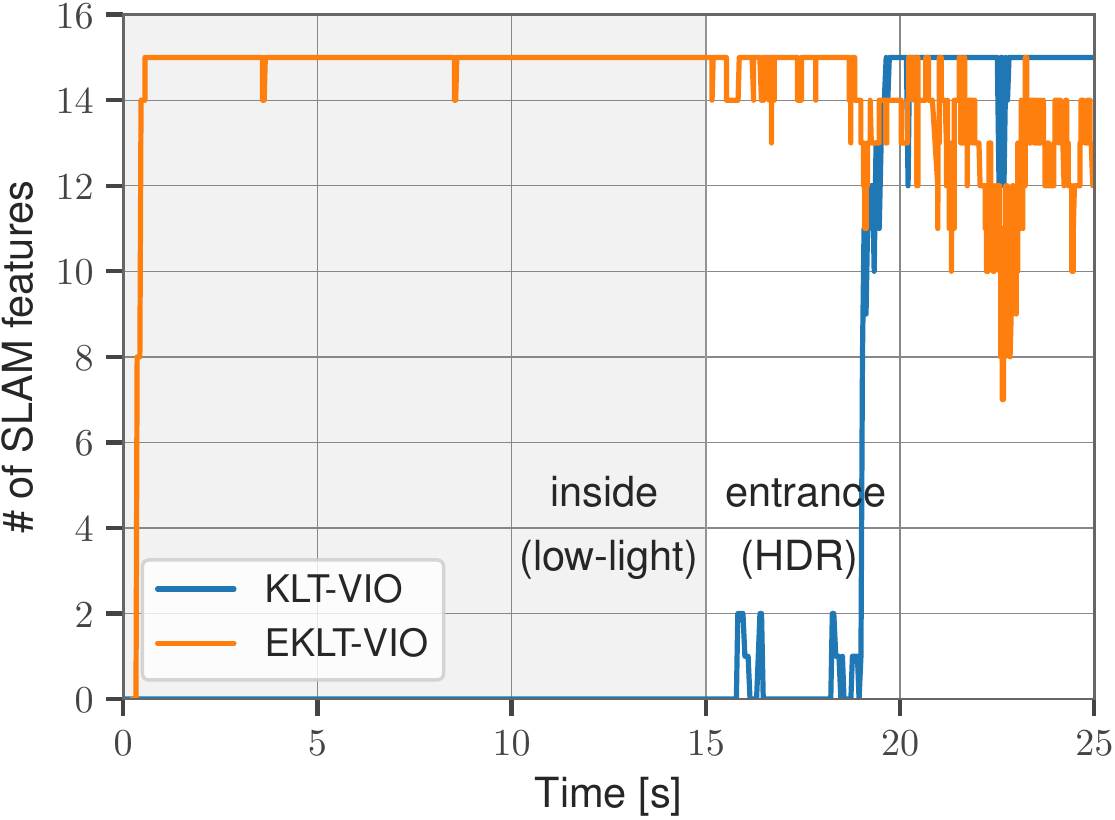}
\end{tabular}
\endgroup
\caption{\revtwo{Tracked features on the Wells Cave sequence. While KLT-VIO and ROVIO quickly diverge, due to lacking features (c), EKLT-VIO can track successfully.}}\label{fig:trajectory_wells_marsyard}\vspace{-2ex}
\end{figure}

We adopt the same evaluation protocol as before and report results for all methods in Tab. \ref{tab:rpg-davis-rotation}.
We observed during this experiment that USLAM did not initialize during these sequences since it could never detect sufficient translation to insert a new keyframe, and it is thus marked with \textit{unfeasible}. 
\revtwo{Frame-based KLT-VIO tracks well for the first 30s, but diverges in the second part, where rapid shaking motion causes motion blur on the frames, and high feature displacements, both of which significantly impact the accuracy of the KLT frontend. This leads to a diverging state estimate.}
By contrast, event-based methods EKLT-VIO and HASTE-VIO can track robustly, because their event-based front-ends are unaffected by motion-blur. EKLT-VIO, however, is the only method to converge on all sequences and yields a consistently lower tracking error compared to all compared methods.
In summary, EKLT-VIO leverages the advantages of event-based frontends for robust high-speed tracking and the advantages of a filter-based backend to fuse small translational motions. This shows that EKLT-VIO is most suitable in these conditions.} 
\vspace{-1ex}
\subsection{Mars-mission Scenario: Wells Cave and JPL Mars yard}
\label{sub:cavern}
\rev{\revtwo{Finally, we show the capabilities of EKLT-VIO in Mars-like exploration scenarios, by comparing it to image-based methods KLT-VIO~\cite{delaune2020xvio}, ORB-SLAM3~\cite{Campos21tro}, OpenVINS~\cite{Geneva20icra}, VINS-Mono~\cite{Qin18tro} and ROVIO~\cite{Bloesch15iros} on sequences recorded at the 
JPL Mars Yard (Fig.~\ref{fig:my_comp} (a)), and Wells Cave Nature Preserve (Fig.~\ref{fig:my_comp} (e)).} The Mars Yard sequence features rapid illumination changes that challenge the autoexposure and result overexposures in the images (Fig.~\ref{fig:my_comp} (b)). The Wells Cave instead is a cave system used by JPL to emulate lava tubes on Mars. It features a low illumination, leading to underexposure in the images (Fig.~\ref{fig:my_comp} (f)). 
In the Wells Cave we use the DAVIS 346\cite{Brandli14ssc}, and in the Mars Yard, we use a mvBlueFOX-MLC200wG standard camera, a DVXplorer event camera, and an MPU9250 IMU. 

\begin{figure*}[t]
\vspace{4ex}
\centering
\begin{tabular}{ccc}
\includegraphics[trim=0 0 0 0,clip,height=0.22\linewidth]{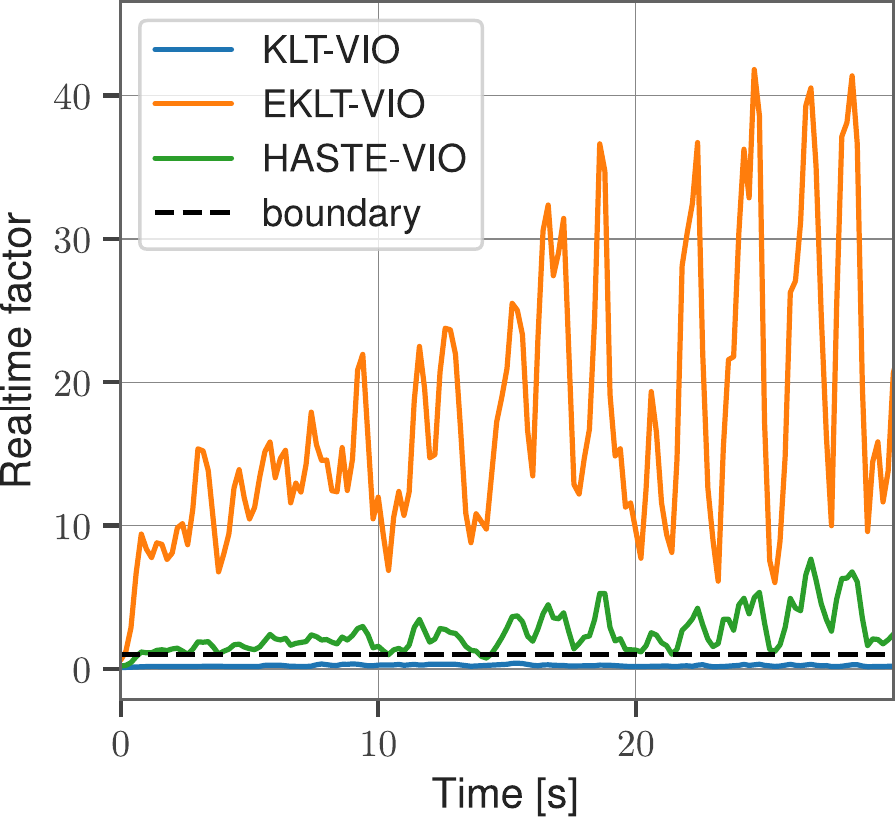}&
\includegraphics[height=0.22\linewidth]{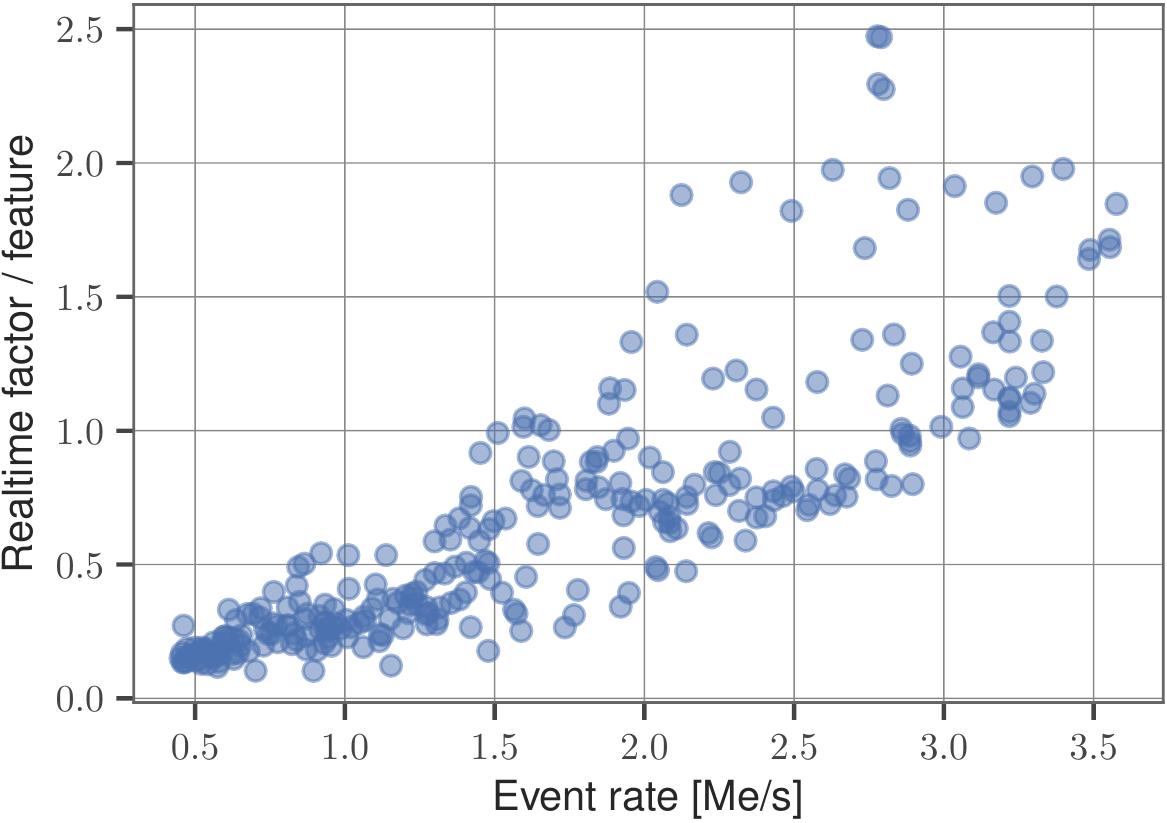}&
\includegraphics[trim=0 -15mm 0 0,clip, height=0.22\linewidth]{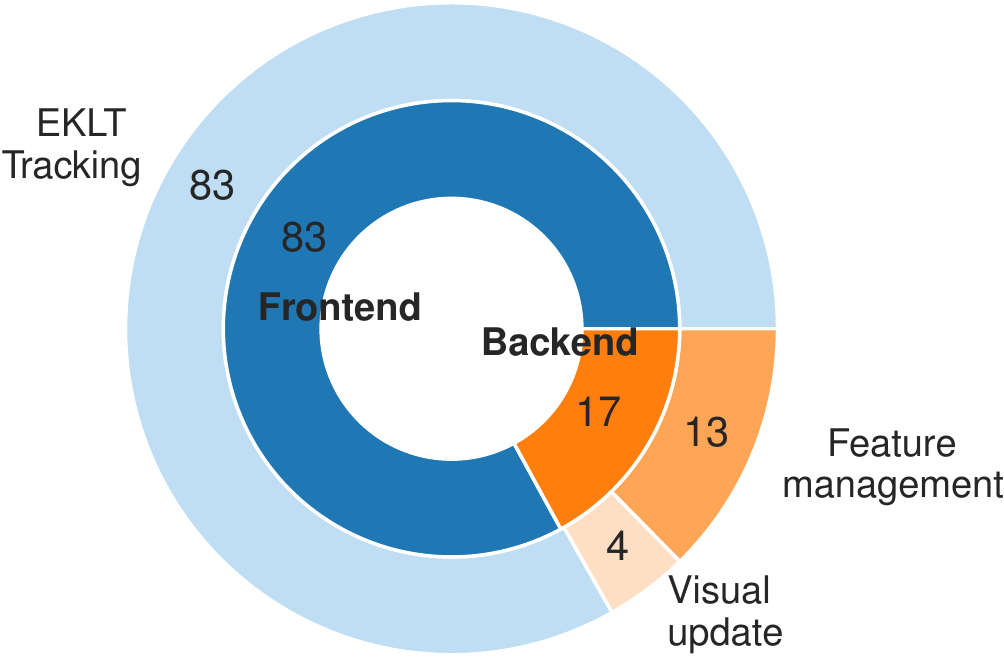}\\
(a) Realtime factor&
(b) Realtime factor vs. Event rate&
(c) Computational pie chart
\end{tabular}
\caption{\rev{Real-time factor (RTF) (a) for EKLT-VIO (orange), HASTE-VIO (green) and KLT-VIO (blue) on \emph{Poster 6DOF}. The RTF per tracked feature (b) increases with the event rate. \revtwo{Our method can process 89'000 events per second when tracking 45 features. As seen in (c), EKLT-VIO spends most of its computation time tracking features.}}}
\label{fig:rt_factor}
\end{figure*}

Here we show that EKLT-VIO can run on events alone, by using images reconstructed from events provided by the method E2VID~\cite{Rebecq19pami}. They feature a much higher dynamic range than the standard images (Fig.~\ref{fig:my_comp} (c,g)). We reconstruct frames every 15'000 events, resulting in an HDR video used by our method. \revtwo{For a resolution of 640 $\times$ 480 these images can be provided with 30 FPS on a Quadro RTX 4000 GPU. However, EKLT-VIO only needs a subset of these images, since it only uses them for feature initialization.}

\textbf{Mars Yard:} The trajectory used in this analysis is a hand-held circular motion with a diameter of 1.5 meters over a sharp shadow with increasing speed. The trajectories tracked by all methods are shown in Fig. \ref{fig:my_comp} (d). \revtwo{While EKLT-VIO consistently tracks the circular motion for at least two revolutions, filter-based methods KLT-VIO and ROVIO diverge due to a lack of features caused by motion blur and HDR conditions. The optimization-based methods ORB-SLAM3 and VINS-MONO fail to initialize, since the sequence starts directly from hover, and misses an initialization trajectory, with which to generate an initial map. \revthree{OpenVINS fails to initialize due to missing parallax.} These methods are therefore not plotted. This shows that thanks to the use of an event-based frontend and filter-based backend EKLT-VIO can overcome this condition.}\\ 
\textbf{Wells Cave:}
Finally, the trajectories in the Wells Cave, for all methods are shown in Fig. \ref{fig:my_comp} (h). \revtwo{Only filter-based methods  KLT-VIO and ROVIO manage to initialize, but diverge quickly. EKLT-VIO tracks consistently, until reaching the tunnel entrance. Again, ORB-SLAM3 and VINS-Mono fail to initialize and therefore are not plotted.}\revthree{OpenVINS fails to initialize due to missing features.} As shown in Fig.~\ref{fig:trajectory_wells_marsyard}, EKLT-VIO consistently maintains SLAM features, while KLT-VIO only does so once it exits the cave.}
\subsection{Limitations}
\rev{
We study EKLT-VIO, KLT-VIO, and HASTE-VIO in terms of their real-time factor (RTF, Fig. \ref{fig:rt_factor} (a)) and report the RTF per feature (b) and computation allocations (c) for EKLT-VIO. We conduct all our experiments on a laptop with an Intel i7-7700HQ quadcore processor\revtwo{, exploiting however only a single core in the current implementation}. The RTF measures how much time is spent to process a second of real-time, and RTF$<1$ indicates real-time performance.    
As seen in Fig.~\ref{fig:rt_factor} (a) there exists a clear speed-accuracy trade-off between EKLT-VIO, HASTE-VIO, and KLT-VIO, since EKLT-VIO achieves a maximum real-time factor of around 45. \revthree{Note that this is 45 times slower than real-time}. For EKLT-VIO, the real-time factor correlates with the event rate (Fig.~\ref{fig:rt_factor} (b)), which depends on the scene texture and camera speed. \revtwo{On \emph{Poster 6DOF} it can process 89'000 kEv/sec.}
}
\revtwo{
\subsection{Speedup Strategies}
Fig.\ref{fig:rt_factor} (c) shows that, the EKLT frontend remains the bottleneck, which directs future work toward speeding up EKLT. 
Tab. \ref{tab:speedup} illustrates three speedup strategies to achieve realtime capabilities evaluated \emph{Poster 6DOF}. (i) We reduce the number of tracked frontend features from 45 to 15, (ii) we increased $n_e$, the number of events before triggering an update, by a factor of two and (iii) we reduce the event rate with random filtering (RF), randomly keeping every $r^{\text{th}}$ event, or refractory period filtering (RPF), where events within a time $\tau$ of the previous event are discarded. To improve the convergence in (ii) we additionally implemented IMU-based feature prediction~\cite{Bloesch15iros}, to improve the initial guess.
While naive RF degrades performance, RPF with $\tau=10$ ms reduces the median RTF to 7.7. Reducing the frontend features results in an RTF of 8.7, and, when combined with filtering, leads to an RTF of 4.2. These steps lead to a minimal increase of the MPE from 0.36 to 0.41. Setting $n_e=6400$ results in an RTF of 9.7, while reducing the MPE from 0.36 to 0.24. However, when combined with additional filtering, we found that the method diverges with an MPE of 3.79, but a lower RTF of 2.05. The remaining gap can be closed by software-side techniques, such as distributing the workload to multiple cores (see \url{https://github.com/Doch88/rpg_eklt_multithreading}). There, up to four cores were parallelized, leading to a 3.6-fold speedup. }



\begin{table}[]
\centering
\resizebox{\linewidth}{!}{%
\begin{tabular}{@{}lrrrr@{}}
\toprule
Speedup method & \multicolumn{1}{l}{MPE} & \multicolumn{1}{l}{MYE} & \multicolumn{1}{r}{RTF Max} & \multicolumn{1}{r}{RTF Median} \\ \midrule
Baseline & 0.36 & 0.02 & 43.6 & 17.9 \\
RF $r=2$ & \multicolumn{2}{c}{\textit{diverging}} & 11.2 & 5.20 \\
RF $r=5$ & \multicolumn{2}{c}{\textit{diverging}} & 5.70 & 2.20 \\
RPF ($\tau=1$ ms) & 0.27 & 0.02 & 37.3 & 15.40 \\
RPF ($\tau=10$ ms) & 0.48 & 0.02 & 15.7 & 7.70 \\
$n_e=6400$ & 0.24 & 0.02 & 21.2 & 9.70 \\
15 Features & 0.31 & 0.02 & 18.6 & 8.70 \\
15 Features, RPF ($\tau=10$ ms)& 0.41 & 0.02 & 8.20 & 4.20 \\ 
15 Features, RPF ($\tau=10$ ms),  $n_e=6400$& 3.79 & 0.02 & 4.39 & 2.05 \\ 
\bottomrule
\end{tabular}%
}
\caption{\revtwo{Real-time factor speedup on \emph{Poster 6DOF}. We compare random filtering (RF), refractory period filtering (RPF), reducing the number of features, and increasing $n_e$. Our baseline tracks 45 features and updates each feature, every $n_e=3200$ events. \revthree{$\text{RTF}>1$ is slower than real-time.}}}
\label{tab:speedup}
\vspace{-4ex}
\end{table}

\section{Conclusion}

Future planetary missions, require us to venture into previously inaccessible domains, such as lava-tubes on Mars, which pose challenging lighting conditions for traditional image-based VIO. We explored the use of event cameras, which promise to shed light in these domains due to their high dynamic range. \rev{We present EKLT-VIO which integrates the state-of-the-art feature tracker EKLT with the filter-based backend xVIO thus leveraging the advantages of both.} The event-based frontend provides robust high-speed feature measurements even in low-light and HDR scenarios while  the filter-based backend addresses the limitations of traditional optimization-based VIO algorithms in near-hovering conditions. 
We show an evaluation on Mars-like sequences and challenging hand-held sequences of the Event-Camera dataset. On these sequences, we demonstrate the robust pose tracking the performance of our methods, showing a mean position error reduction of up to 32\% compared to event- and frame-based state-of-the-art methods. 
Additionally, we showcase the advantages of our backend and frontend in the first successful evaluation on the rotation-only sequences of the Event-Camera Dataset with fast motion and challenging lighting conditions.
Finally, we demonstrate our method's robustness in visually challenging conditions recorded in the JPL Mars Yard \rev{and in the Wells Cave, replicating our mission scenario.} 
To spur further research in this direction, we open-source the implementation of this work and release our Mars-like sequences.






\bibliographystyle{IEEEtran}
\bibliography{RA-L-Template-YY.bbl}

\end{document}